\documentclass[conference]{IEEEtran}
\IEEEoverridecommandlockouts
\usepackage{cite}
\usepackage{amsmath,amssymb,amsfonts}
\usepackage{graphicx}
\usepackage{textcomp}
\usepackage{xcolor}
\usepackage{cleveref}
\usepackage{amsmath}
\usepackage{algorithm}
\usepackage{algpseudocode}

\usepackage{multirow}
\usepackage{tabularx}
\usepackage{booktabs} 

\usepackage[]{hyperref}

\hypersetup{
hidelinks,
colorlinks=true,
linkcolor=black,
citecolor=black
}

\def\BibTeX{{\rm B\kern-.05em{\sc i\kern-.025em b}\kern-.08em
    T\kern-.1667em\lower.7ex\hbox{E}\kern-.125emX}}
\begin{document}

\title{LDCA: Local Descriptors with Contextual Augmentation for Few-Shot Learning
\\
{\footnotesize \textsuperscript{*}}
\thanks{}
}

\author{Maofa Wang,
Bingchen Yan \\
1321847667a@gmail.com,\\Guilin University Of Electronic Technology}

\maketitle

\begin{abstract}
Few-shot image classification has emerged as a key challenge in the field of computer vision, highlighting the capability to rapidly adapt to new tasks with minimal labeled data. Existing methods predominantly rely on image-level features or local descriptors, often overlooking the holistic context surrounding these descriptors. In this work, we introduce a novel approach termed “Local Descriptor with Contextual Augmentation (LDCA)”. Specifically, this method bridges the gap between local and global understanding uniquely by leveraging an adaptive global contextual enhancement module. This module incorporates a visual transformer, endowing local descriptors with contextual awareness capabilities, ranging from broad global perspectives to intricate surrounding nuances. By doing so, LDCA transcends traditional descriptor-based approaches, ensuring each local feature is interpreted within its larger visual narrative. Extensive experiments underscore the efficacy of our method, showing a maximal absolute improvement of 20\% over the next-best on fine-grained classification datasets, thus demonstrating significant advancements in few-shot classification tasks. Additionally, our approach significantly elevates the quality of local descriptors, minimizing traditional k-nearest neighbor classification model (k-NN) sensitivity to the choice of k, especially in scenarios with scarce training samples. We posit that the LDCA framework paves the way for a new paradigm in few-shot learning, where local features are augmented with rich contextual insights for enhanced discriminative power.
\end{abstract}

\section{Introduction}
\label{sec:intro}

Few-shot learning distinguishes itself from most contemporary artificial intelligence algorithms by eschewing the dependency on high-quality, large-scale training datasets. Instead, it employs a limited number of supervised samples to train deep learning models, aiming to emulate human-like abilities to learn new knowledge quickly from imperfect data. This approach holds significant importance in applying deep learning models to areas where acquiring large-scale, high-quality datasets is challenging, such as medical image processing, remote sensing image scene classification, and biomedical relationship extraction.
Presently, popular methods in few-shot learning are broadly categorized into two types: task-level optimization methods \cite{finn2017model,cai2018memory,jamal2019task,santoro2016meta,zhmoginov2022hypertransformer,rusu2018meta,baik2020meta,lee2019meta} and metric-based methods \cite{snell2017prototypical,huang2022compound,zhang2022kernel,li2019revisiting,qiao2019transductive,li2020more,zheng2023bdla,song2023learning,vinyals2016matching}. A notable advancement in metric-based methods is the introduction of the local descriptor-based image-to-class module by Li\cite{li2019revisiting}. This module addresses the issue where summarizing an image's local features into a compact image-level representation may result in the loss of irrevocable discriminative information. Furthermore, it introduces a method for calculating the image-to-class metric by performing k-nearest neighbor searches on local descriptors, which has led to significant improvements.

\begin{figure*}[ht]
\centering
\includegraphics[scale=0.7]{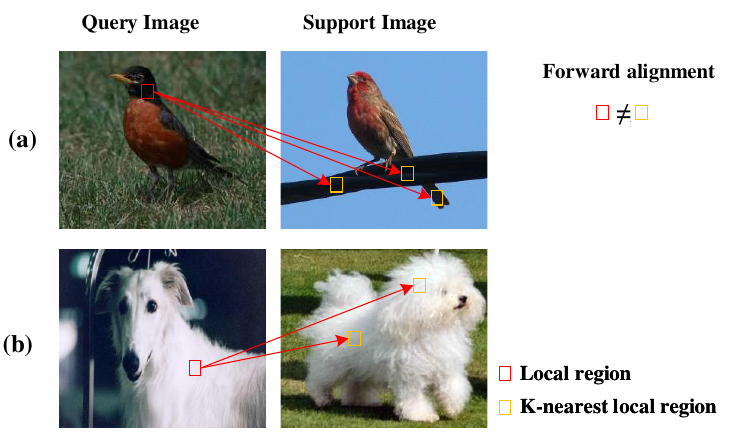}
\caption{(a)The illustration of samples belongs to the same class.Feature misalignment occurs when the local descriptor of the query (highlighted in red) mistakenly associates with a similarly colored but irrelevant background region in the support image (enclosed in yellow). This is indicative of the limitations inherent in methods that rely solely on direct feature comparison without contextual consideration.(b)The illustration showcases samples from different classes and highlights the challenge of distinguishing ambiguous regions within fine-grained classification datasets. These datasets frequently contain repetitive patterns, such as texture, color, and shape, which complicate differentiation when relying solely on local information}
\label{figl}
\end{figure*}

The prevalent use of Convolutional Neural Networks (CNNs) as feature extractors in almost all current few-shot models, due to their inherent locality \cite{d2021convit}, limits feature extraction to a finite receptive field, neglecting semantic and spatial information beyond the local area. This constraint leads to two potential drawbacks. First, as shown in \cref{figl}(a), semantic misalignment can occur when the dominant object in a query sample resembles the background information of a support sample \cite{zheng2023bdla}. To address this, BDLA \cite{zheng2023bdla} builds on the DN4 framework, proposing the calculation of bidirectional distances between query and support samples to enhance effective alignment of contextual semantic information. DLDA \cite{song2023learning} suggests assigning weights based on the ratio of intra-class to inter-class similarity for each local descriptor by finding its k-nearest neighbors. However, both methods still consider context within a fixed regional neighborhood. The second drawback, illustrated in \cref{figl}(b), is the challenge of differentiating ambiguous areas in fine-grained classification datasets, which often feature repetitive patterns (including texture, color, shape, etc.) using only local information, leading to modest improvements in recent algorithms on such datasets.

As shown in \cref{main}.Our proposed LDCA model leverages the robust capabilities of the visual transformer architecture to amalgamate local descriptors with global context, thereby enriching the information contained within local descriptors and enhancing their representational power. Subsequent experiments demonstrate that the enhanced local descriptors exhibit exceptional performance, especially in fine-grained classification datasets, achieving inspiring improvement results. Furthermore, the enhanced local descriptors reduce the traditional $k$-NN classification model's sensitivity to the choice of $k$, mitigating fluctuations in accuracy due to varying $k$-value selections in the $k$-NN model.

\section{Related Works}
{\bf Few-shot Learning.}
Few-shot learning aims to train deep neural networks with a limited amount of labeled data and extend the acquired knowledge to new classes that also have few labeled samples. Current image classification algorithms based on few-shot learning fall into two main categories: task-level optimization methods and metric-based methods.

Task-level optimization methods, also known as meta-learning approaches like MAML \cite{finn2017model}, MetaOptNet\cite{lee2019meta}etc., focus on learning a good initialization that serves as knowledge and experience. This allows for rapid adaptation to new tasks through one or multiple gradient update steps, rather than starting from scratch each time. Metric-based methods primarily classify or regress by learning a function that measures the similarity between samples. For instance, Prototypical Networks\cite{snell2017prototypical} compute a 'prototype' (the average representation) for each class in the feature space. A new sample is classified based on its distance to all prototypes, with the nearest prototype determining its predicted class. Matching Networks \cite{cai2018memory} calculate the similarity of each sample in the support set to a query sample and predict the query sample's label based on a weighted sum of these similarities.

In this paper, we focus on metric-based methods. As previously mentioned, Li \cite{li2019revisiting}. innovatively utilized metric at the local descriptor level, which alleviates the issue of losing substantial discriminative information that might occur when summarizing an image's local features into a compact image-level representation. They proposed using an image-to-class approach, calculating k-nearest local features from the local descriptors of each query example to the support examples, achieving significant results. BDLA \cite{zheng2023bdla} introduced the computation of bidirectional distances between query and support samples to strengthen the effective alignment of contextual semantic information. DLDA \cite{song2023learning} and MADN4 \cite{li2020more} proposed weighting each local descriptor to reduce the impact of noise, thereby obtaining more representative local descriptors.

{\bf Context-awareness.}
Context awareness plays a pivotal role in a wide array of computer vision tasks, including image classification \cite{wang2018non,zhang2018recursive,chen20182}, image semantic segmentation \cite{zhang2018context,zhang2019co,yu2020context,yang2021context}, instance segmentation \cite{bolya2019yolact,chen2019hybrid}, object detection \cite{zhang2020feature,wang2020deep}, and person re-identification \cite{yan2018participation,tang2020blockmix}. To achieve context awareness, one approach, as exemplified by ASPP \cite{chen2017deeplab}, PPM \cite{zhao2017pyramid}, and MPM \cite{hou2020strip}, involves defining a larger receptive field through deeper architectures to aggregate multi-scale context based on spatially adjacent pixels. Another approach includes non-local interactions \cite{zhang2020feature,wang2018non}, self-attention \cite{zhang2019co,vaswani2017attention}, and  object context \cite{yuan2020object,zhang2019positional}, where each feature location participates in the global context computation, facilitating context awareness based on remote dependency relationships.

Recently, the visual transformer (ViT) \cite{wu2020visual} has demonstrated its capability to aggregate global context by segmenting input images into 16×16 tokens using patch embedding and directly applying the transformer architecture to visual tasks, as evidenced in \cite{carion2020end,wang2020axial}. However, the majority of ViT applications today focus on pixel-level tasks, with its application on the level of local descriptors not yet widely recognized. Inspired by Wang . \cite{wang2023attention}, who applied ViT to local descriptors and achieved encouraging results in image matching, homography estimation, visual localization, and 3D reconstruction tasks, our proposed LDCA  model integrates the visual transformer with a learnable gating map. This integration adaptively embeds global context and positional information into local descriptors, reducing the intrinsic loss of image feature information while ensuring the extraction of potentially representative data. Our experiments demonstrate that the enhanced local descriptors, due to increased distinctiveness, significantly improve the model's ability to differentiate ambiguous areas in fine-grained classification datasets with repetitive patterns (including texture, color, shape, etc.). Compared to the DN4 model, our LDCA model reduces the sensitivity of the $k$-NN classifier to the choice of $k$-values.

\section{The Proposed Method}
\subsection{Problem Definition}
Few-shot learning primarily investigates how to enable models to learn with very few samples while achieving robust generalization. Specifically, we focus on the $M$-way $K$-shot problem, where $M$ denotes the number of classes, and $K$ represents the number of samples in each class. Typically, the value of $K$ is small, such as 1 or 5.

Given a training set \(D^{train} = \{(x_r, y_r)\}^T_{r=1}\), our goal is to learn model parameters $\theta$, enabling rapid adaptation to an unseen test set \(D^{test}\) within an episodic training mechanism \cite{vinyals2016matching}. Each \(y_r\) represents the true label of the image \(x_r\). In \(D^{train}\) and \(D^{test}\), each episode contains a support set \(S\) and a query set \(Q\). The support set \(S\) consists of $M$ different image classes, with $K$ randomly labeled images in each class. The query set \(Q\) is used for model evaluation.

These datasets comprise three parts: training, validation, and testing sets. The label space of each part does not overlap with the others, meaning the classes seen during the training phase will not reappear in the validation or testing phases. Typically, each part contains more classes and samples than $M$ and $K$. Each part can be divided into multiple episodes, each containing a support set and a query set, randomly drawn from the corresponding dataset. Notably, the support set and query set are non-overlapping, but they share the same label space.

To simulate real-world few-shot learning scenarios, all training, validation, and testing procedures are based on this episodic mechanism. During training, the model randomly selects an episode in each iteration for parameter updates, a process repeated multiple times until the model converges to a stable state. In the validation and testing phases, the obtained model is used to classify the query set \(Q\) based on the support set \(S\).

\subsection{Framework of the Proposed Method}
\begin{figure*}[ht]
\centering
\includegraphics[scale=0.9]{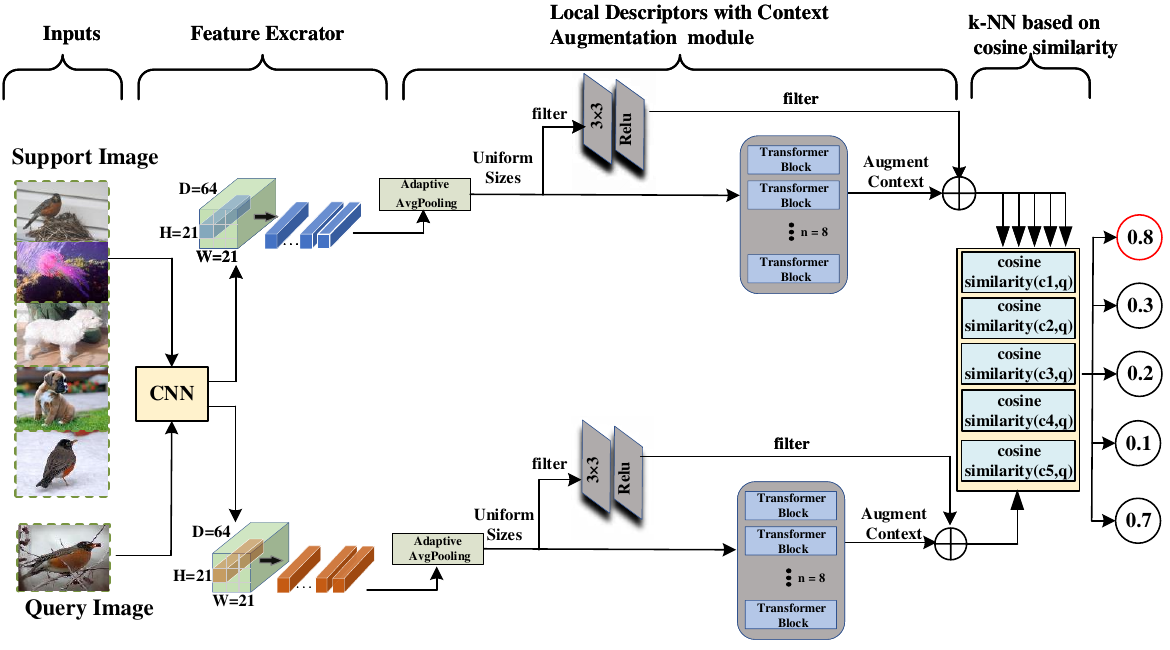}
\caption{The proposed LDCA method's framework for 5-way 5-shot classification consists of three key components: (i) a feature embedding model, utilizing a CNN to extract local descriptors from images; (ii) a contextual augmentation model that adaptively integrates global context and positional information into the local descriptors of both support and query images; (iii) a k-NN based classifier that computes the similarity between query set images and each class in the support set.c$i$ represents the $i$-th class, $q$ represents the query set image}
\label{main}
\end{figure*}

As illustrated in \cref{main}, our approach comprises three primary components: a feature extractor, a context-enhanced local descriptor model, and a classification model. In keeping with conventional practice, the feature embedding model is composed of a Convolutional Neural Network (CNN), utilized for extracting features from images in both the support and query sets. This process results in deep local descriptors for all images. To achieve local descriptors imbued with global contextual information, the extracted local descriptors are augmented through our newly proposed Local Descriptor with Contextual Augmentation (LDCA) module.
In the final classification stage of our model, we integrate a $k$-Nearest Neighbors ($k$-NN) algorithm to evaluate the similarity between a query image and each class represented in the support set. This is achieved by calculating the distance between each local descriptor of the query image and the descriptors within each class. For every local descriptor, we identify its $k$-nearest neighbors, measure the cosine similarity with these neighbors, and then aggregate these measurements. Specifically, we sum the cosine similarities across all  spatial locations and the top \( k \) neighbors to yield a comprehensive similarity score for each class. The query image is then assigned to the class with the highest cumulative similarity score, thus leveraging both local and neighborhood information to inform the classification decision.

\subsection{Feature Embedding Model}

By passing each image \( X \) through the Image Embedding model, we obtain a 3D(three-dimensional) tensor \( \mathcal{F}_\theta(X) \in \mathbf{R}^{D \times H \times W} \).

This represents the image, where \( \mathcal{F}_\theta(X) \) is the hypothesized function learned by the neural network, \( \theta \) represents the parameters of the neural network, and \( D \), \( H \), \( W \) denote the depth, height, and width of the 3D tensor, respectively. This can be expressed as:

\begin{equation}
    \mathcal{F}_\theta(X) = \left[\boldsymbol{x}^1, \ldots, \boldsymbol{x}^M\right] \in \mathbf{R}^{D \times M}
\end{equation}

Here, \( M = H \times W \), mapping all images to a representational space. Each 3D tensor contains \( M \) units of \( D \) dimensions, where each unit represents a local descriptor of the image. Compared to 1D \cite{vinyals2016matching,snell2017prototypical} or other dimensional representations, 3D tensors capture geometric information more effectively. Therefore, in few-shot learning within metric learning, 3D tensors are a more common choice. In this paper, we employ three-dimensional tensor features to represent the corresponding support set \( S \) and query set \( Q \).

\subsection{Local Descriptors with Context Augmentation Model}
In this study, we introduce a novel model, termed the Local Descriptor Contextual Augmentation (LDCA) Model. Our method fundamentally addresses two main issues: First, traditional CNNs, when dealing with a small number of samples, tend to extract features only within a local receptive field, overlooking the broader context's semantic and spatial information, leading to potential semantic mismatches \cite{zheng2023bdla}. Second, they struggle with ambiguous areas in fine-grained datasets characterized by repetitive patterns. To counter these challenges, the LDCA model integrates the visual transformer architecture to enhance global contextual information in local descriptors.

Specifically, consider an image $X_s$ from the support set S and an image $X_q$ from the query set Q. The output of the feature extractor $\mathcal{F}\theta(X)$ yields 3D tensors representing the local descriptors of $X_s$ as $\mathcal{F}\theta(X) = \left[\boldsymbol{x}^1_s, \ldots, \boldsymbol{x}^M_s\right] \in \mathbf{R}^{D \times M}$ and those of $X_q$ as $\mathcal{F}_\theta(X) = \left[\boldsymbol{x}^1_q, \ldots, \boldsymbol{x}^M_q\right] \in \mathbf{R}^{D \times M}$.Initially, we reshape the 3D tensors into a 64×64×64 size using adaptive average pooling. For fine-grained image classification tasks, higher resolution feature maps provide more accurate spatial information. Following \cite{wu2020visual}, the 3D tensor is reshaped from 64×64×64 into a sequence of flattened 2D patches $x_p$, each with a shape of 16×16. The Transformer maintains a constant latent vector size of 128 across all its layers. Consequently, the patches are flattened and mapped to 128 dimensions with a trainable linear projection. This projection's output serves as the patch embeddings.
 
 To endow local descriptors with relative global spatial positioning information, a learnable embedding is added to the sequence of the embedded patches to preserve positional information, as shown below:
\begin{equation}
\begin{aligned}
    \mathbf{z}_0 & =\left[ \mathbf{x}_p^1 \mathbf{E}; \mathbf{x}_p^2 \mathbf{E}; \cdots; \mathbf{x}_p^N \mathbf{E}\right]+\mathbf{E}_{\text{pos}}
\end{aligned}
\end{equation}

where ${E}$ represents the patch embedding projection and $E_{pos}$ represents the positional embedding. After passing through eight transformer blocks, ${z}_0$ yields hidden features enriched with global context. The transformer layer structure, as shown in  supplementary material, consists of alternating layers of Multihead Self-Attention (MSA) and MLP blocks.

Specifically, the output of the $k^{th}$ layer can be expressed as:
\begin{equation}
    \begin{aligned}
    \mathbf{z}_{k}^{\prime} & =\operatorname{MSA}\left(\operatorname{LN}\left(\mathbf{z}_{k-1}\right)\right)+\mathbf{z}_{k-1}, & & k=1 \ldots K \\
    \mathbf{z}_{k} & =\operatorname{MLP}\left(\operatorname{LN}\left(\mathbf{z}_{k}^{\prime}\right)\right)+\mathbf{z}_{k}^{\prime}, & & k=1 \ldots K
    \end{aligned}
\end{equation}
where MSA(·) represents multihead self-attention \cite{wu2020visual}, MLP(·) denotes a multilayer perceptron block, and LN(·) signifies layer normalization. Layernorm (LN) is applied before every block, and residual connections after every block. The MLP includes two layers with a GELU non-linearity.

After reshaping, we obtain the globally context-enhanced patch descriptors:
\begin{equation}
    \mathbf{y} = \operatorname{LN}\left(\mathbf{z}_K^0\right)
\end{equation}
Furthermore, to empower regions where local information is insufficient for effective description—such as areas where the dominant object in the query sample resembles the background information of the support sample, and repetitive patterns in fine-grained classification datasets—with a global perspective on spatial information, thereby enhancing their distinctiveness, we adhere to the idea of weighting local descriptors to increase discriminative power, as demonstrated in \cite{li2020more,song2023learning}. A ReLU activation function is employed to implement a gating mechanism for filtering local descriptors. Finally, we combine the gated, filtered local descriptors with the globally context-enhanced patch descriptors to enhance the model's ability to recognize repetitive patterns and ambiguous areas in fine-grained datasets.


With this unique design, the LDCA model significantly improves recognition performance in various complex scenarios, showcasing its powerful generalization and adaptability.

\begin{algorithm}
\caption{Context-Enhanced Local Descriptor Classification}
\begin{algorithmic}[1]
\Require Support set images $S$, Query set images $Q$, $k$ for k-NN
\Ensure Class label for each image in $Q$

\State // Feature Embedding Model
\For{each image $X$ in $S \cup Q$}
    \State Compute 3D tensor $F_\theta(X) \in \mathbb{R}^{D \times H \times W}$
    \State Reshape to $F_\theta(X) \in \mathbb{R}^{D \times M}$ where $M = H \times W$
\EndFor

\State // Local Descriptors with Context Augmentation (LDCA) Model
\For{each image $X$ in $S \cup Q$}
    \State Reshape 3D tensor to 64×64
    \State Flatten to sequence of 2D patches $x_p$
    \State Project patches to $D$ dimensions: $z_0 = [x_p^1 \mathbf{E}; x_p^2 \mathbf{E}; \ldots; x_p^N \mathbf{E}] + \mathbf{E}_{\text{pos}}$
    \For{$k = 1$ to $K$}
        \State $z_k' = \text{MSA}(\text{LN}(z_{k-1})) + z_{k-1}$
        \State $z_k = \text{MLP}(\text{LN}(z_k')) + z_k'$
    \EndFor
    \State Compute globally context-enhanced descriptors $y = \text{LN}(z_K)$
    \State Apply ReLU for gating and filtering
    \State Combine with original descriptors to enhance discriminative power
\EndFor

\State // Classification using k-NN
\For{each query image $X_q$ in $Q$}
    \For{each class $c$ in $S$}
        \State Calculate distance between descriptors of $X_q$ and $c$
        \State Identify k-nearest neighbors and compute cosine similarity
        \State Aggregate similarities to form class score for $c$
    \EndFor
    \State Assign $X_q$ to class with highest score
\EndFor

\end{algorithmic}
\end{algorithm}
\subsection{High-Quality Local Descriptors Reduce Classifier Sensitivity to the Choice of $k$}
The final stage involves classification, for which numerous methods can be employed to realize the classifier's functionality. In this study, we continue the approach of \cite{li2019revisiting,li2020more,zheng2023bdla,song2023learning} by employing a $k$-Nearest Neighbors ($k$-NN) model based on cosine similarity as the classifier. A notable drawback of this method is its implicit assumption that the $k$-nearest neighbors are equally important in the classification decision, regardless of their distance from the query point. This assumption can lead to significant fluctuations in the final classification results in few-shot classification problems due to the impact of different $k$ values. To alleviate this issue, the approach of \cite{zheng2023bdla} follows \cite{li2019revisiting} in enumerating model accuracies under different $k$ values and selecting the optimal $k$ value for different tasks. The method in \cite{song2023learning} assigns different weights to each nearest neighbor based on their distance from the query point to increase the discriminative power of different neighbors.

To investigate whether the fluctuations in final model accuracy due to different $k$-value selections are attributable to the mere lack of distinctiveness in local descriptors, we conducted multiple comparative experiments with several advanced models. Additionally, we employed High Dimensional Discriminant Analysis/Clustering models \cite{bouveyron2007high} to cluster local descriptors before and after enhancement with LDCA. It was observed that the integration of global contextual information into local descriptors led to improved feature representation and classification performance, as evidenced by the e Supplementary' figures.

\section{Experiments}
\subsection{Datasets}

\textbf{MiniImageNet.} This subset \cite{vinyals2016matching}, derived from ImageNet \cite{vinyals2016matching}, includes 100 classes with 600 images each. Each image has a resolution of 84x84 pixels. Following the methodology of \cite{ravi2016optimization}, these classes are divided into 64 for training, 16 for validation, and 20 for testing.

\textbf{CUB-200.} This is a fine-grained bird image classification dataset \cite{welinder2010caltech} involving 200 different bird species. The number of images varies between classes. 130 classes are used for training, 20 for validation, and the remaining 50 for testing.

\textbf{Stanford Dogs.} This dataset \cite{khosla2011novel} focuses on fine-grained dog image classification, comprising 20,580 photographs across 120 different dog breeds. Images of 70 breeds are used for training, 20 breeds for validation, and the remaining 30 breeds for testing.

\textbf{Stanford Cars.} This dataset \cite{krause20133d} is designed for fine-grained car image classification, containing 16,185 images across 196 different car classes, defined based on brand, model, and year of manufacture. 130 classes are used for training, 17 for validation, and the remaining 49 for testing.

To maintain consistency, all images in CUB-200, Stanford Dogs, and Stanford Cars have been resized to 84x84 pixels, matching the image size of MiniImageNet.
\subsection{Experimental Setting}
In our experiments, we focused primarily on 5-way 1-shot and 5-shot classification tasks. During the training phase, we employed episodic training mechanisms, constructing numerous task sets from the training portions of each dataset. For each training task, we selected $K$ support images per class, along with 15 and 10 query images for 1-shot and 5-shot settings, respectively. For instance, in a 5-way 1-shot task, each training episode would include 5 support images and a total of 75 query images. To train our model, we utilized the Adam optimization algorithm \cite{kingma2014adam} with an initial learning rate of 0.001. We randomly sampled and constructed 300,000 episodes for training all our models, halving the learning rate every 100,000 episodes.

For the testing phase, we randomly constructed 600 episodes from the test portions of each dataset to evaluate the model's performance. The number of episodes in both training and testing phases was determined experimentally and aligned with the settings used in other methods, ensuring fairness in our experiments. To ensure the reliability of our results, we repeated the testing process multiple times and calculated the average accuracy along with a 95\% confidence interval. It is noteworthy that our model was trained entirely from scratch in an end-to-end manner, with no fine-tuning during the testing phase.

Besides, we maintained consistency with established few-shot learning methodologies by adopting the network design structure commonly used in metric-learning based approaches. This ensures a fair comparison with other methods as referenced in \cite{li2019revisiting,li2020more,zheng2023bdla,song2023learning}. Our feature embedding model, named Conv4 for its simplicity, consists of four convolutional blocks. Each block comprises a 3×3 convolutional layer with 64 filters, followed by a batch normalization layer and a LeakyReLU activation function. To efficiently manage the output size, we incorporated a 2×2 max-pooling layer at the end of the initial two convolutional blocks, aligning with configurations used in previous works. This uniform network structure allows for direct and equitable comparisons across different few-shot learning models within our experimental framework.
\begin{table*}
    \centering
    \begin{tabular}{l|l|c|c}
        \toprule
        \textbf{Model} & \textbf{Embedding} & \multicolumn{2}{c}{\textbf{5-Way Accuracy (\%)}} \\
        \cmidrule{3-4}
        & & \textbf{1-shot} & \textbf{5-shot} \\
        \midrule
        Ren . \cite{ren2018meta} & Conv4 & 50.41$\pm$0.31 & 64.39$\pm$0.24 \\
        Reptile \cite{nichol2018first} & Conv4 & 49.97$\pm$0.32 & 65.99$\pm$0.58 \\
        Ravichandran . \cite{ravichandran2019few} & Conv4 & 49.07$\pm$0.43 & 65.73$\pm$0.36 \\
        ML-LSTM \cite{ravi2016optimization} & Conv4-32 & 43.44$\pm$0.77 & 60.60$\pm$0.71 \\
        MAML \cite{finn2017model} & Conv4-32 & 48.70$\pm$1.84 & 63.11$\pm$0.92 \\
        MetaGAN \cite{zhang2018metagan} & Conv4-32 & 52.71$\pm$0.64 & 68.63$\pm$0.67 \\
        Matching Net \cite{vinyals2016matching} & Conv4-64 & 43.56$\pm$0.84 & 55.31$\pm$0.73 \\
        Prototype Net \cite{snell2017prototypical} & Conv4-64 & 49.42$\pm$0.78 & 68.20$\pm$0.66 \\
        GNN \cite{garcia2017few} & Conv4-64 & 49.02$\pm$0.98 & 63.50$\pm$0.84 \\
        Relation Net \cite{sung2018learning} & Conv4-64 & 50.44$\pm$0.82 & 65.32$\pm$0.70 \\
        PABN \cite{huang2019compare} & Conv4-64 & 51.87$\pm$0.45 & 65.37$\pm$0.68 \\
        TPN-semi \cite{liu2018learning} & Conv4-64 & 52.78$\pm$0.27 & 66.42$\pm$0.21 \\
        DN4 \cite{li2019revisiting} & Conv4-64 & 51.24$\pm$0.74 & 71.02$\pm$0.64 \\
        mAP-SSVM \cite{triantafillou2017few} & Conv4-64 & 50.32$\pm$0.80 & 63.94$\pm$0.72 \\
        Meta-SGD \cite{li2017meta} & Conv4-64 & 50.47$\pm$1.87 & 64.03$\pm$0.94 \\
        R2-D2 \cite{bertinetto2018meta} & Conv4-512 & 51.80$\pm$0.20 & 68.40$\pm$0.20 \\
        BDLA \cite{zheng2023bdla} & Conv4-64 & 52.97$\pm$0.35 & 71.31$\pm$0.68 \\
        MADN4 \cite{li2020more} & Conv4-64 & 53.20$\pm$0.52 & 71.66 $\pm$0.47 \\
        DLDA \cite{song2023learning} & Conv4-64 &  52.81 $\pm$0.79 & 71.76  $\pm$0.66\\
        our LDCA(k=1) & Conv4-64 &  53.03 $\pm$0.63 & 74.02  $\pm$0.49\\
        our LDCA(k=3) & Conv4-64 &  \textbf{53.46} $\pm$\textbf{0.63} & \textbf{75.06}  $\pm$\textbf{0.48}\\

        \bottomrule
    \end{tabular}
    \caption{Comparison with state-of-the-art methods in the 5-way 1-shot and 5-shot settings: Average classification accuracies (\%) are provided for  the MiniImageNet dataset, along with 95\% confidence intervals. The experiments are conducted using the Conv4 network to ensure a fair comparison}
    \label{tab:results1}
\end{table*}

Furthermore, to substantiate the effectiveness of our approach, we conducted comparative analyses with several existing methods. For the MiniImageNet dataset, we compared our method with the following 12 approaches: MAML \cite{finn2017model}, TAML \cite{jamal2019task}, MetaLearner LSTM \cite{ravi2016optimization}, MetaGAN \cite{zhang2018metagan}, GNN \cite{garcia2017few}, TPN-semi \cite{liu1805transductive}, Relation Net \cite{sung2018learning}, Matching Net \cite{vinyals2016matching}, Prototypical Net \cite{snell2017prototypical}, DN4 \cite{li2019revisiting}, BDLA \cite{zheng2023bdla}, and MADN4 \cite{li2020more}. For three fine-grained datasets, we compared our approach with five methods: Matching Net \cite{vinyals2016matching}, Prototypical Net \cite{snell2017prototypical}, DN4 \cite{li2019revisiting}, BDLA \cite{zheng2023bdla}, and GNN \cite{garcia2017few}. These comparisons were designed to showcase the performance and advantages of our method across various datasets.

\subsection{Comparison with state-of-the-art models}
By conducting experimental comparisons with several models on four benchmark datasets, including a general MiniImageNet dataset and three few-shot fine-grained datasets, we have validated the effectiveness and superiority of the proposed LDCA model.

\textbf{Comparative experimental results on MiniImageNet dataset.}
\cref{tab:results1} presents the experimental comparison results of the state-of-the-art (SOTA) methods on the miniImagenet \cite{vinyals2016matching} for 1-shot and 5-shot settings, along with 95\% confidence intervals. Note that in Table 1, Conv4-n indicates a 4-layer convolutional network producing feature maps with n channels. The comparison results from Table 1 reveal that our LDCA method outperforms most of the previous metric-based methods in both 5-way 1-shot and 5-way 5-shot classification settings. The comparisons with methods that directly weight local descriptors, such as DLDA and MADN4, further demonstrate the superiority of our approach of first enriching local descriptor information to increase their discriminability, followed by weighted classification.

\textbf{Comparative experimental results on Fine-grained Datasets.}
Building on the preset conditions of the MiniImageNet dataset, we further conducted experiments on three major fine-grained image datasets: Stanford Dogs \cite{khosla2011novel}, Stanford Cars \cite{krause20133d}, and CUB-200 \cite{welinder2010caltech}. Typically, fine-grained datasets present greater intra-class variance and smaller inter-class differences, making them more challenging for few-shot classification compared to traditional classification tasks.
\begin{table*}
    \centering
    \begin{tabular}{|l|l|c|c|c|c|c|c|}
        \hline
        \multirow{2}{*}{Model} & \multirow{2}{*}{Embedding} & \multicolumn{2}{c|}{Stanford Dogs} & \multicolumn{2}{c|}{Stanford Cars} & \multicolumn{2}{c|}{CUB-200} \\
        \cline{3-8}
         &  & 1-shot & 5-shot & 1-shot & 5-shot & 1-shot & 5-shot \\
        \hline
        PCM \cite{wei2019piecewise} & Conv4-64 & 28.78$\pm$2.33 & 46.92$\pm$2.00 & - & - & 42.10$\pm$1.96 & 62.48$\pm$1.21 \\
        Matching Net \cite{vinyals2016matching} & Conv4-64 & 35.80$\pm$0.99 & 47.50$\pm$1.03 & 34.80$\pm$0.98 & 44.70$\pm$1.03 & 45.30$\pm$1.03 & 59.50$\pm$1.01 \\
        Prototype Net \cite{snell2017prototypical} & Conv4-64 & 37.59$\pm$1.00 & 48.19$\pm$1.03 & 40.90$\pm$1.01 & 52.93$\pm$1.03 & 37.36$\pm$1.00 & 45.28$\pm$1.03 \\
        GNN \cite{garcia2017few} & Conv4-64 & 46.98$\pm$0.98 & 62.27$\pm$0.95 & 55.85$\pm$0.97 & 71.25$\pm$0.89 & 51.83$\pm$0.98 & 63.69$\pm$0.94 \\
        DN4 \cite{li2019revisiting} & Conv4-64 & 45.41$\pm$0.76 & 63.51$\pm$0.62 & 59.84$\pm$0.80 & 88.65$\pm$0.44 & 46.84$\pm$0.81 & 74.92$\pm$0.62 \\
        BDLA \cite{zheng2023bdla} & Conv4-64 & 48.53$\pm$0.87 & 70.07$\pm$0.70 & \textbf{64.41}$\pm$\textbf{0.84} & 89.04$\pm$0.45 & 50.59$\pm$0.97 & 75.36$\pm$0.72 \\
        DLDA \cite{song2023learning} & Conv4-64 & 49.44$\pm$0.85 & 69.36$\pm$0.69 & 60.86$\pm$0.82 & 89.50$\pm$0.41 & 55.12$\pm$0.86 & 74.46$\pm$0.65 \\
        MADN4 \cite{li2020more} & Conv4-64 & 50.42$\pm$0.27 & 70.75$\pm$0.47 & 62.89$\pm$0.50 & 89.25$\pm$0.34 & 57.11$\pm$0.70 & 77.83$\pm$0.40 \\
        our LDCA(k=1) & Conv4-64 & \textbf{55.72}$\pm$\textbf{0.72} & \textbf{80.76}$\pm$\textbf{0.48} & 56.80$\pm$0.66 & \textbf{91.91}$\pm$\textbf{0.34} & \textbf{75.45}$\pm$\textbf{0.67} & \textbf{91.63}$\pm$\textbf{0.33} \\

        \hline
    \end{tabular}
    \caption{Comparison with state-of-the-art methods in the 5-way 1-shot and 5-shot settings: Average classification accuracies (\%) are provided for the fine-grained datasets, along with 95\% confidence intervals. The experiments are conducted using the Conv4 network to ensure a fair comparison}
    \label{tab:results2}
\end{table*}

As shown in \cref{tab:results2},in the 5-way 1-shot setting, our LDCA method exhibited excellent performance on the Stanford Dogs and CUB-200 datasets. Compared to the DN4 method, which does not process local descriptors, our LDCA achieved a 10.31\% gain on the Stanford Dogs dataset and a 28.61\% gain on the CUB-200 dataset. Against methods that directly weight local descriptors, such as DLDA and MADN4, our LDCA gained 6.28\% and 5.3\% on the Stanford Dogs dataset, and 20.33\% and 18.34\% on the CUB-200 dataset, respectively. In the 5-way 5-shot scenario, our LDCA significantly outperformed methods that directly weight local descriptors, such as DLDA, MADN4, and BDLA, which employs a bidirectional local alignment strategy, on both the Stanford Dogs and CUB-200 fine-grained datasets.


\subsection{Transfer Learning on Fine-grained Datasets}
The primary distinction between fine-grained datasets and conventional image datasets lies in their focus on subtle differences in images. The main objective of cross-domain classification is to transfer knowledge acquired in the source domain to the target domain, thereby validating the model's generalization performance in the target domain. The fine-grained datasets, with their focus on minute image variations, aptly meet this requirement. Therefore, to verify the adaptability of our model, we used the MiniImageNet dataset as the source domain for training and Stanford Dogs \cite{khosla2011novel}, Stanford Cars \cite{krause20133d}, and CUB-200 \cite{welinder2010caltech} datasets as target domains for testing.

We compared our proposed method with three methods that do not optimize local descriptors \cite{li2019revisiting,zheng2023bdla,song2023learning} as shown in \cref{results3}. As indicated in \cref{results3}, in the 5-way 1-shot classification task, the LDCA method was slightly outperformed by DLDA on the Stanford Dogs dataset. This experimental result suggests that models using context-enhanced local descriptors can achieve better transferability. It further substantiates the superiority of our approach, which first enriches local descriptor information to increase their discriminability and then employs weighted classification.

\subsection{Ablation analyses}
The $k$-nearest neighbors model, serving as a classifier, is used to align the similar semantic information between local descriptor features of images. To explore the impact of different $k $values on the LDCA model's results and to demonstrate that enhanced local descriptors increase the stability of the traditional $k$-NN classifier compared to non-enhanced ones, we conducted a parameter analysis of k-nearest neighbors using the DN4, BDLA, and our LDCA models on the MiniImageNet benchmark dataset. Specifically, we compared our LDCA method with BDLA and DN4, which do not modify $k$-NN, choosing different $k$-nearest neighbors (i.e., $k$ = {1,3,5,7}) for experimentation. We put the related experimental results in the supplementary
material. As shown in the supplementary material, our proposed LDCA model exhibits significantly reduced sensitivity to the value of $k$ in both 5-way 1-shot and 5-way 5-shot settings compared to the models proposed by the first two methods. This further demonstrates that the discriminability of local descriptors is notably enhanced by using context augmentation to enrich local descriptor information, which is the reason for the improved classification accuracy of the subsequent classifiers.

Furthermore, regarding the reshaping of 3D tensors into a size of 64×64 through adaptive average pooling, as discussed in Section 3.4, we note that for fine-grained image classification tasks, feature maps of higher resolution provide more accurate spatial information. Accordingly, we have provided experimental comparisons in the supplementary material.


\begin{table*}
\begin{tabular}{lllllll}
\hline 
Dataset & & Proto Net \cite{snell2017prototypical} & Relation Net \cite{sung2018learning} & BDLA \cite{zheng2023bdla}  & DLDA \cite{song2023learning}& LDCA (ours) \\
\hline 
Stanford Dogs & 5-way 1-shot & $33.11 \pm 0.64$ & $31.59 \pm 0.65$ & $35.55 \pm 0.66$ & $\textbf{37.10} \pm \textbf{0.70}$ & $36.53 \pm 0.52$ \\
& 5-way 5-shot & $45.94 \pm 0.65$ & $41.95 \pm 0.62$ & $52.64 \pm 0.69$ & $53.99 \pm 0.70$ & $\textbf{56.92} \pm \textbf{0.56}$ \\
Stanford Cars & 5-way 1-shot & $29.10 \pm 0.75$ & $28.46 \pm 0.56$ & $30.62 \pm 0.58$ & $31.48 \pm 0.56$ & $\textbf{32.68} \pm \textbf{0.48}$ \\
& 5-way 5-shot & $38.12 \pm 0.60$ & $39.88 \pm 0.63$ & $45.99 \pm 0.61$ & $49.63 \pm 0.66$ & $\textbf{52.69} \pm \textbf{0.55}$ \\
CUB-200 & 5-way 1-shot & $39.39 \pm 0.68$ & $39.30 \pm 0.66$ & $40.40 \pm 0.76$ & $41.36 \pm 0.74$ & $\textbf{45.98} \pm \textbf{0.59}$ \\
& 5-way 5-shot & $56.06 \pm 0.66$ & $53.44 \pm 0.64$ & $58.23 \pm 0.72$ & $60.02 \pm 0.71$ & $\textbf{68.03} \pm \textbf{0.53}$ \\
\hline
\end{tabular}
\caption{The average accuracy of the 5-way 1-shot and 5-way 5-shot accuracy on the different fined-grained datasets by training model on the
MiniImageNet dataset, with 95\% confidence intervals}
\label{results3}
\end{table*}

\section{Conclusions}
In conclusion, our research contributes a novel perspective to the field of few-shot learning by introducing the Local Descriptor with Contextual Augmentation (LDCA) model. This model synergizes the strengths of visual transformer architecture with conventional Convolutional Neural Networks, thereby enriching local descriptors with global contextual information. This integration not only enhances the representational power of these descriptors but also addresses critical limitations in existing few-shot learning approaches, such as semantic misalignment and challenges in fine-grained classification. The LDCA model's efficacy is particularly notable in datasets with intricate patterns, where it demonstrates superior performance compared to existing models. Moreover, our approach mitigates the sensitivity of the traditional $k$-NN classification model to the choice of \( k \), thereby offering more stable and reliable classification results.  Our findings not only underscore the potential of integrating local and global contextual information in deep learning models but also pave the way for future advancements in this rapidly evolving field.

\bibliographystyle{elsarticle-num}
\bibliography{reference}

\end{document}